\documentclass[runningheads]{llncs}
\usepackage{graphicx}
\usepackage{siunitx}
\usepackage{algorithm}
\usepackage{algpseudocode}
\usepackage{xcolor}
\usepackage{amsmath,amssymb} 
\usepackage{bm}      
\usepackage{booktabs}
\usepackage{multirow}
\usepackage{subcaption}
\usepackage{hyperref}

\usepackage{soul}
\usepackage{enumitem}
\begin{document}
%
\title{Sparse Double Descent in Vision Transformers: real or phantom threat?}
\titlerunning{}
%
\author{Victor Qu\'etu\orcidID{0009-0004-2795-3749} \and Marta Milovanovi\'c\orcidID{0000-0002-3280-2396} \and Enzo Tartaglione\orcidID{0000-0003-4274-8298}} 

\authorrunning{V. Qu\'etu et al.}
%
\institute{LTCI, T\'el\'ecom Paris, Institut Polytechnique de Paris, France
\email{\{name.surname\}@telecom-paris.fr}}
\maketitle              
\begin{abstract}
Vision transformers (ViT) have been of broad interest in recent theoretical and empirical works. They are state-of-the-art thanks to their attention-based approach, which boosts the identification of key features and patterns within images thanks to the capability of avoiding inductive bias, resulting in highly accurate image analysis.
Meanwhile, neoteric studies have reported a ``sparse double descent'' phenomenon that can occur in modern deep-learning models, where extremely over-parametrized models can generalize well.
This raises practical questions about the optimal size of the model and the quest over finding the best trade-off between sparsity and performance is launched: are Vision Transformers also prone to sparse double descent? Can we find a way to avoid such a phenomenon?\\
Our work tackles the occurrence of sparse double descent on ViTs. Despite some works that have shown that traditional architectures, like Resnet, are condemned to the sparse double descent phenomenon, for ViTs we observe that an optimally-tuned $\ell_2$ regularization relieves such a phenomenon. However, everything comes at a cost: optimal lambda will sacrifice the potential compression of the ViT.

\keywords{Sparse double descent, transformers, pruning, deep learning}
\end{abstract}
\section{Introduction}
\let\svthefootnote\thefootnote
\newcommand\freefootnote[1]{%
  \let\thefootnote\relax%
  \footnotetext{#1}%
  \let\thefootnote\svthefootnote%
}
Deep\freefootnote{Article accepted for publication at the 22nd International Conference on Image Analysis and Processing (ICIAP23).} neural networks (DNNs) have revolutionized the field of computer vision by achieving state-of-the-art results in tasks such as segmentation~\cite{chaudhry2022lung}, classification~\cite{barbano2022two}, and object detection~\cite{mazzeo2022image}. DNNs outperform conventional machine learning algorithms on many visual recognition tasks as they can automatically learn feature representations from raw input~\cite{dosovitskiy2021an}. In addition, they can process a lot of data and generalize well to novel, unseen examples. For a long time, convolutional neural architectures (CNN) like VGG and ResNet models have been dominant in computer vision, thanks to their ability to learn, simultaneously, feature extraction (typically handled by convolutional layers) and classification (by multi-layer perceptrons, or in some cases even by convolutional layers themselves, like in ALL-CNN~\cite{SpringenbergDBR14}).\\ 
A new, deep architecture called Transformer has been pioneered by~\cite{vaswani2017attention}, and it has, at first, been conceived for natural language processing tasks~\cite{brown2020language}, resulting in a break-through for the community. Given its big potential, the computer vision world has recently begun to adopt it~\cite{Liu2021SwinTH}. Vision Transformers (ViT), which are Transformer architectures adapted for computer vision, quickly became state-of-the-art for many tasks, out of which we cite generative models~\cite{esser2021taming}. However, the lack of strong inductive biases causes them to be even more data-hungry than traditional CNN architectures, and this poses severe performance drops when noisy data are available to train such a model.\\
In image classification tasks, noisy labels are a frequent issue that can negatively impact the performance of deep learning models~\cite{sukhbaatar2014training}. Incorrect labels in the training data can mislead the model during the learning process and result in sub-optimal performance. Various approaches have been proposed to address this issue, including label smoothing, data augmentation, and robust loss functions~\cite{ma2020normalized}. The expected behavior, in such cases, is that the higher the noise in the data, the higher the overfit the model will suffer. As opposed to the traditional bias-variance trade-off, a phenomenon has been recently discovered, called Double Descent (DD)~\cite{Nakkiran2021Deep}. Namely, enlarging the model size in the over-fitting regime worsens the performance of an over-parametrized network; then, the trend reverses. DD represents an important challenge in finding the optimal set of parameters since it shows that it is possible to potentially improve generalization in an over-parametrized regime, but without real indicators on the best model's size to adopt. This behavior is observed in various architectures, stretching from machine learning models to DNNs, such as standard CNNs and ResNet~\cite{yilmaz2022regularization}. Analogously, a sparse double descent (SDD) phenomenon is observed when moving the model from an over-parametrized towards a sparser regime~\cite{SparseDoubleDescent}, via parameter pruning. \\
In this paper, we show that ViTs also suffer from the SDD phenomenon: besides the burden of the lack of an inductive bias that could help these models to generalize, the occurrence of SDD makes the performance even worse in intermediate compression regimes, when only a part of the parameters is removed. This is a possible explanation for the fact that typical ViT architectures can not be pruned to similar extreme rates as traditional CNNs~\cite{yu2022width}. However, contrarily to what is suggested for traditional CNNs~\cite{quetu2023dodging}, ViTs can avoid SDD with the optimal tuning of $\ell_2$ regularization, a result which was suggested by the theory~\cite{nakkiran2021optimal}. We postulate this is possible thanks to the lack of an inductive bias embedded in the architecture, which favors the strong regularization necessary to avoid SDD. Such a discovery enables back all the traditional compression mechanisms, having as a stop criterion a worsening in performance on the validation set. Everything, however, comes at a cost: we observe that optimally regularized models are significantly less compressible, due to the strong prior we impose to avoid SDD. 
We summarize, here below, our key messages and contributions.
\begin{itemize}[nolistsep, noitemsep, topsep=-\parskip]
    \item To the best of our knowledge, this is the first paper raising concerns on the potential occurrence of the sparse double descent phenomenon in ViT models. Through this work, we compare the behavior of ViT and ResNet in the typically employed test scenarios~\cite{Nakkiran2021Deep}, also including a test on real annotated data (CIFAR-100N), observing SDD also on ViT.
    \item We propose a quantitative study over the $\ell_2$ regularization parameter, supported by the theory~\cite{nakkiran2021optimal} but already proven as inapplicable to traditional CNNs~\cite{quétu2023avoid}. We observe that, in ViT models, avoiding SDD is possible with a properly tuned value for the regularization, which nicely imposes a strong prior on the model's parameters, impossible in traditional architectures suffering from inductive bias.
    \item We interestingly observe a trade-off between avoidance of SDD and compressibility of the model. More specifically, to avoid SDD (to employ typical pruning schemes with a stop criterion once the performance on a validation set worsens below some given threshold) we want to have a strong $\ell_2$ regularization, which however makes the model less compressible as a higher number of parameters will have a similar relevance. Depending on what we are targeting (high performance or high compressibility) we might want or not want to avoid SDD.
\end{itemize}
\section{Background on Vision Transformers}
\begin{figure*}[t]
    \centering
    \begin{subfigure}{0.6\textwidth}
        \includegraphics[width=\textwidth]{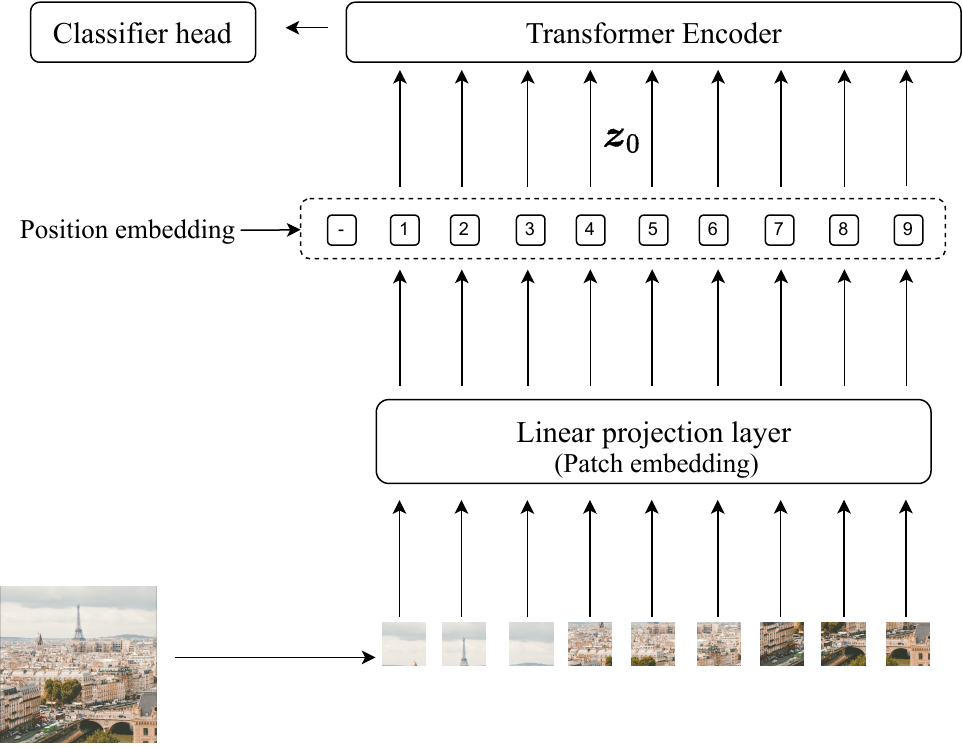}
        \caption{~}
        \label{ViT}
    \end{subfigure}
    \begin{subfigure}{0.3\textwidth}
        \includegraphics[width=\textwidth]{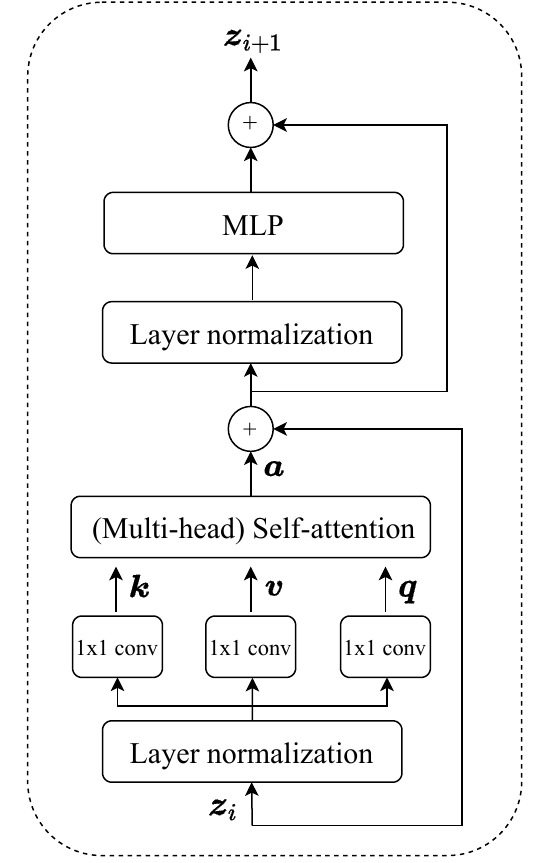}
        \caption{~}
        \label{Transformer_encoder}
    \end{subfigure}
    \caption{Vision Transformer (ViT) (a) and Transformer Encoder (b).}
    \label{fig:Transformer}
\end{figure*}

Vision Transformers~\cite{DosovitskiyB0WZ21}, presented in Fig.~\ref{ViT}, use self-attention mechanisms to capture the relationships between the elements of an input image. They essentially consist of four key elements: patch embedding, positional encoding, the transformer encoder, shown in Fig.~\ref{Transformer_encoder}, and the classification head.\\
\noindent\textbf{Patch embedding.} The input image is divided into a grid of patches (which can or can not be non-overlapping), each containing a fixed number of pixels. These patches are then linearly projected to a lower-dimensional embedding space. This process converts the spatial information of the image into a sequence of patch embeddings, mimicking the process of text embedding. \\
\noindent\textbf{Positional encoding.} To preserve the positional information of the image patches, positional encoding is added to the patch embeddings: this allows to distinguish different patches and capture their global positions in the image.
\\
\noindent\textbf{The transformer encoder.} The patch embeddings, along with their positional encodings, are fed into a stack of transformer encoder layers. Each encoder layer consists of two sub-layers: a multi-head self-attention mechanism, and a multi-layer perception (MLP). The self-attention mechanism enables the model to capture global interactions between patches by attending to all patches and aggregating information accordingly, thus performing feature aggregation. More specifically, self-attention operations determine the attention output $\boldsymbol{a}$ based on the relevance of one item to others. This is iteratively refined and computed using keys $\boldsymbol{k}$ and queries $\boldsymbol{q}$, which have the same dimension \emph{d}, and values $\boldsymbol{v}$~\cite{vaswani2017attention}. The keys are the indices of the hidden states of the encoded input items, and each key $k_i$ has some associated value $v_i$. Each query $q_i$ represents an output coming from the encoded target item (class). The attention is computed as the softmax of the product between $\boldsymbol{q}$ and $\boldsymbol{k}$, and then multiplied by $\boldsymbol{v}$: the model learns to prioritize important input features and capture more informative representations of the input data. Many attention heads are hence concatenated, to form the multi-head, to obtain contextualized representations that include both local and global information. Finally, an MLP carries out feature transformation.\\
\noindent \textbf{The classification head.} At the end of the transformer encoder stack, a classification head is attached to the output of the final transformer layer. The classification head can take various forms, such as a simple fully connected layer or a combination of linear and softmax layers. It maps the aggregated representation of the patches to class probabilities, in order to perform image classification.\\
\noindent \textbf{Training.} ViT models require pretraining on enormous datasets (such as JFT-300M, consisting of approximately 300 million images) due to their lack of strong inductive bias, which is present in other architectures like CNNs~\cite{dehghani2023scaling}. Despite recent advances in learning on smaller datasets using distillation approaches or optimizing models with smaller sizes, transformers still have larger model architectures compared to CNN-based models and require large datasets for optimal performance, despite requiring less computational resources~\cite{DosovitskiyB0WZ21}. However, \cite{transformers_in_vision_survey2022} concludes that scaling up Transformer models improves performance, but with current designs, it is computationally expensive and necessitates efficient designs.\\
\noindent \textbf{Beyond traditional ViT.} In the last few years, many different Vision Transformer designs have been proposed to improve the performance of computer vision tasks. One of the most popular is SWIN~\cite{Liu2021SwinTH} which proposes shifted windows to create overlapping receptive fields, cascaded stages to mimic a multi-resolution approach, tokenization of windows, and token shifts across the stages. As it is possible to imagine, this architecture already goes in the direction of customizing the Transformer architecture to process images. Other newly proposed transformers variants include, among others, CoaT~\cite{dai2021coatnet}, TNT~\cite{yuan2021tokens}, and DeiT~\cite{touvron2021training}.

\section{Sparse Double Descent and ViT}
\begin{figure}[t]
    \centering
    \includegraphics[width=0.9\linewidth]{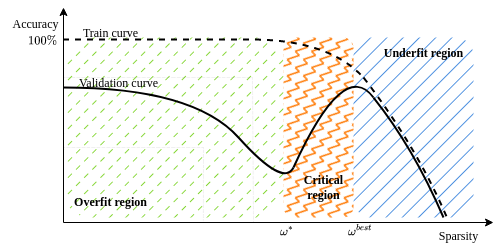}
    \caption{The Sparse Double Descent phenomenon.}
    \label{fig:SDD}
\end{figure}
\begin{algorithm}[t]
\caption{Iterative algorithm to detect Sparse Double Descent.}
\label{Algo}
\begin{algorithmic}[1]
\Procedure{DETECT\_SDD ($\boldsymbol{w}^{\text{init}}$, $\Xi$, $\lambda$, $\zeta^{\text{iter}}$,$\zeta^{\text{end}}$)}{}
\State $\boldsymbol{w} \gets$ Train($\boldsymbol{w}^{\text{init}}$, $\Xi^{\text{train}}$, $\lambda$)\label{line:dense}
\State $p_{i-1} \gets $ Performance($\boldsymbol{w}$,$\Xi^{\text{val}}$)
\State prev\_increasing, prev\_decreasing, already\_increased, already\_decreased $\gets$ False
\State SDD $\gets$ False
\While{Sparsity($\boldsymbol{w}, \boldsymbol{w}^{\text{init}}$) $< \zeta^{end}$}\label{line:endcond}
\State $\boldsymbol{w} \gets$ Prune($\boldsymbol{w}$, $\zeta^{\text{iter}}$) \label{line:prune}
\State $\boldsymbol{w} \gets$ Train($\boldsymbol{w}$,$\Xi^{\text{train}}$, $\lambda$)\label{line:wd} 
\State $p_i \gets $ Performance($\boldsymbol{w}$,$\Xi^{\text{val}}$)
\If {($p_i < p_{i-1}$ {\bf and} already\_decreased {\bf and not} prev\_decreasing) {\bf or}\\~~~~~~~~~~~~($p_i > p_{i-1}$ {\bf and} already\_increased {\bf and not} prev\_increasing)}
    \State SDD $\gets$ True
\EndIf
\If {$p_i \neq p_{i-1}$}
\State prev\_decreasing $\gets p_i < p_{i-1}$; prev\_increasing $\gets p_i > p_{i-1}$
\State already\_decreasing $\gets$already\_decreasing {\bf or} $ p_i < p_{i-1}$
\State already\_increasing $\gets$already\_increasing {\bf or} $ p_i > p_{i-1}$
\State $ p_i \gets p_{i-1}$
\EndIf
\EndWhile
\State {\bf Return }SDD
\EndProcedure
\end{algorithmic}
\end{algorithm}
In this section, we will discuss the background for Double Descent and Sparse Double Descent, moving then to the potential impact on ViT architectures.

\noindent \textbf{Double descent.} It is known that when comparing a model performance (on unseen data) and model complexity, as the complexity grows (from right to left), we observe a first region where the performance improves (under-fitting - blue region in Fig.~\ref{fig:SDD}) and then, at some point, a trend inversion where the performance decreases while increasing the model's complexity (over-fitting). When exposed to real-world noisy data, however, neural networks tend to exhibit the DD phenomenon~\cite{Nakkiran2021Deep}: instead of being monotonous in that region, the performance inverts, at some point, its trend (critical region - orange in Fig.~\ref{fig:SDD}), and starts back decreasing (overfit region - green in Fig.~\ref{fig:SDD}).\\  
DD has been observed in regression tasks and successfully averted with optimally-tuned $\ell_2$ regularization~\cite{nakkiran2021optimal}. However, for classification tasks, this problem is not easily mitigated. It has been shown that the more challenging the dataset and classification task, the harder it is to avoid DD~\cite{quétu2023avoid}. The authors in~\cite{Nakkiran2021Deep} demonstrate the DD not only depending on the model width but also depending on the number of epochs during training. Similarly to DD, an SDD phenomenon happens in the transition from the complex model toward the sparse, pruned model (as illustrated in Fig.~\ref{fig:SDD})~\cite{SparseDoubleDescent}. SDD has implications for model selection, regularization techniques, and understanding the behavior of complex models in high-dimensional settings, as the presence of SDD makes many criteria, like when to stop the pruning, unclear.\\
\noindent\textbf{Addressing the Sparse Double Descent}. We introduce here Alg.~\ref{Algo}, designed to demonstrate the eventual occurrence of the sparse double descent phenomenon. The algorithm begins by training the model on the learning task $\Xi$ for the first time, incorporating $\ell_2$ regularization weighted by $\lambda$ (line~\ref{line:dense}). Following this initial training step, a magnitude pruning stage is set up (line~\ref{line:prune}). Neural network pruning aims to reduce the size of a large network while maintaining its accuracy by removing irrelevant weights, filters, or other structures. As in \cite{SparseDoubleDescent}, we use in this algorithm an unstructured pruning method called magnitude-based pruning, popularized by~\cite{han2015learning}, in which a fixed amount of weights below some specific threshold, are pruned (line~\ref{line:prune}). Here, every time we prune, a fixed $\zeta^{\text{iter}}$ fraction of parameters from the model is removed.
We highlight that more complex pruning approaches exist, but magnitude-based pruning shows its competitiveness despite very low complexity~\cite{Gale_Magnitude}. The accuracy of the model typically decreases after pruning. To improve the performance of the model, we retrain it using the same original learning policy (line~\ref{line:wd}). Recent works have shown that this approach leads to the best performance at the highest sparsities~\cite{quetu2023dodging}. This approach allows us to determine whether a sparsely-parameterized model, starting from its initialization, has the potential to successfully learn a given target task. We end our pruning procedure once we reach a sparsity $\zeta^{\text{end}}$ (line~\ref{line:endcond}).

\noindent\textbf{ViT and Sparse Double Descent.} The number of parameters in ViT architectures is proportional to the model depth and quadratic function of the width. There is a tendency to scale these models even further, to increase their performance~\cite{dehghani2023scaling}, even though this is becoming very computationally expensive. Looking from that perspective, the understanding of the comportment of the ViT models becomes essential. Having a completely different learning architecture from other models, like CNNs, it is not easy to predict the behavior of ViT when pruning is applied. Our work addresses this issue and performs an extensive study with different levels of label noise and various model sparsity levels. In the next section, we will conduct a quantitative study on ViT, determining whether SDD is a real threat to ViT as it is to CNNs or not.

\section{Experiments}
\begin{figure*}[t]
    \begin{subfigure}{0.5\textwidth}
        \includegraphics[width=\textwidth]{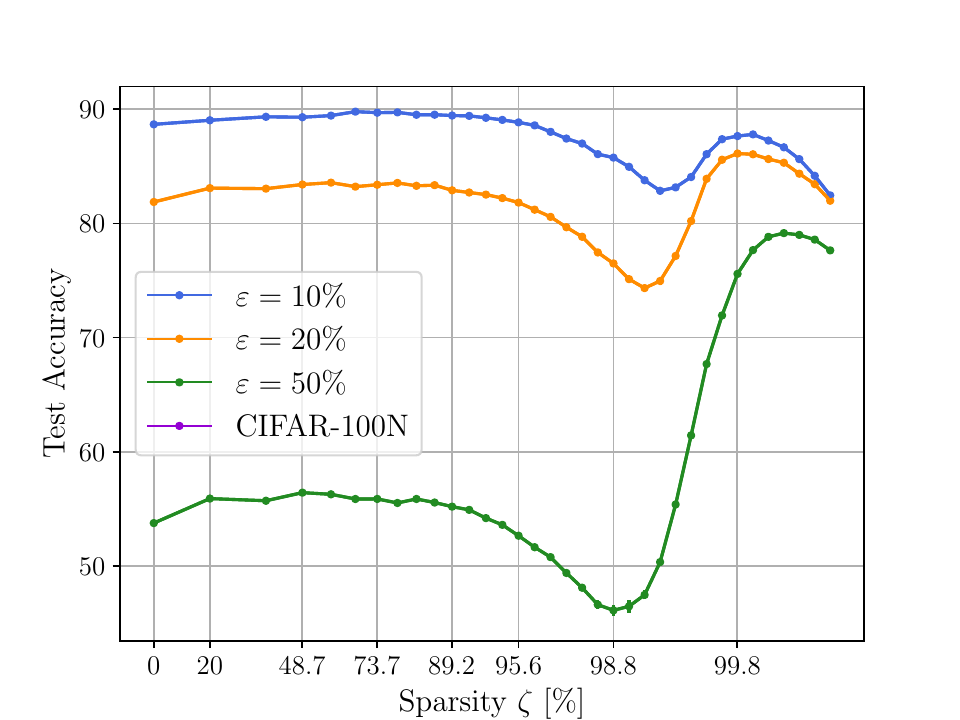}
        \caption{ResNet-18 on CIFAR-10}
        \label{fig:C10-R18}
    \end{subfigure}
    \begin{subfigure}{0.5\textwidth}
        \includegraphics[width=\textwidth]{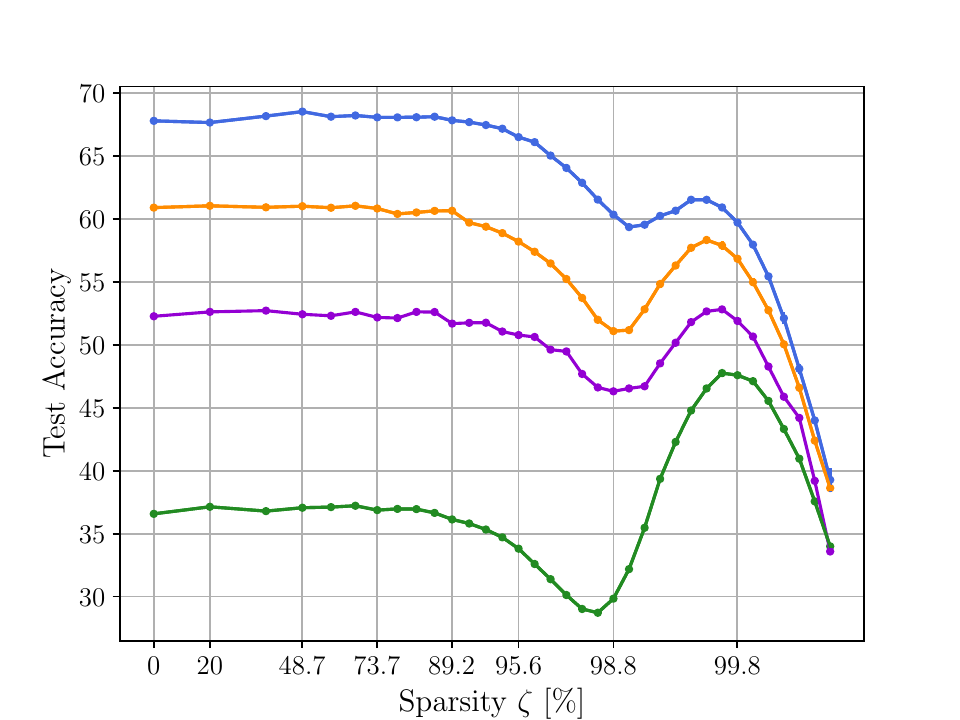}
        \caption{ResNet-18 on CIFAR-100}
        \label{fig:C100-R18}
    \end{subfigure}
    \label{fig:CIFAR-R18}
    \begin{subfigure}{0.5\textwidth}
        \includegraphics[width=\textwidth]{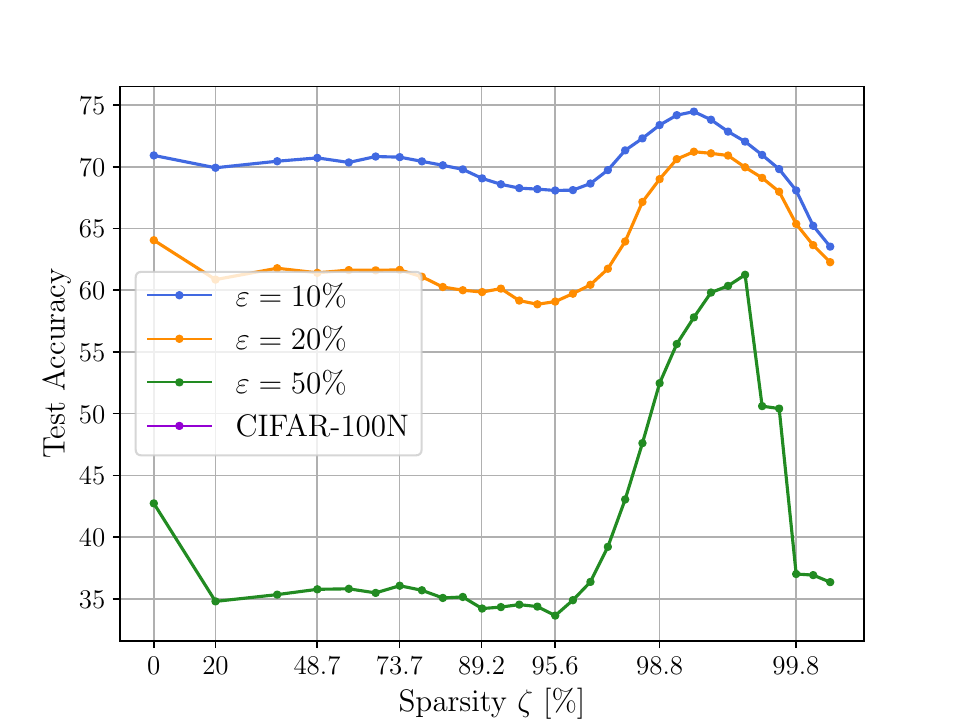}
        \caption{ViT on CIFAR-10}
        \label{fig:C10-ViT}
    \end{subfigure}
    \begin{subfigure}{0.5\textwidth}
        \includegraphics[width=\textwidth]{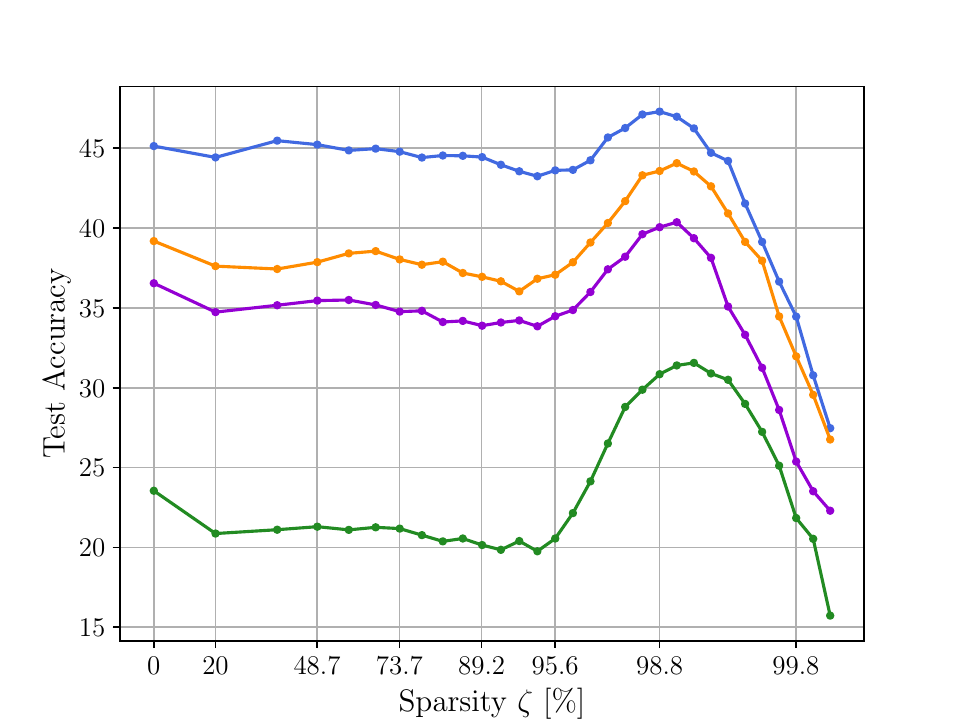}
        \caption{ViT on CIFAR-100}
        \label{fig:C100-ViT}
    \end{subfigure}
    \caption{Test accuracy of ResNet-18 and ViT on CIFAR dataset with different amount of noise $\varepsilon$.}
    \label{fig:CIFAR-ViT}
    \vspace{-10pt}
\end{figure*}
\begin{figure}[t]
    \centering
    \includegraphics[width=0.7\linewidth]{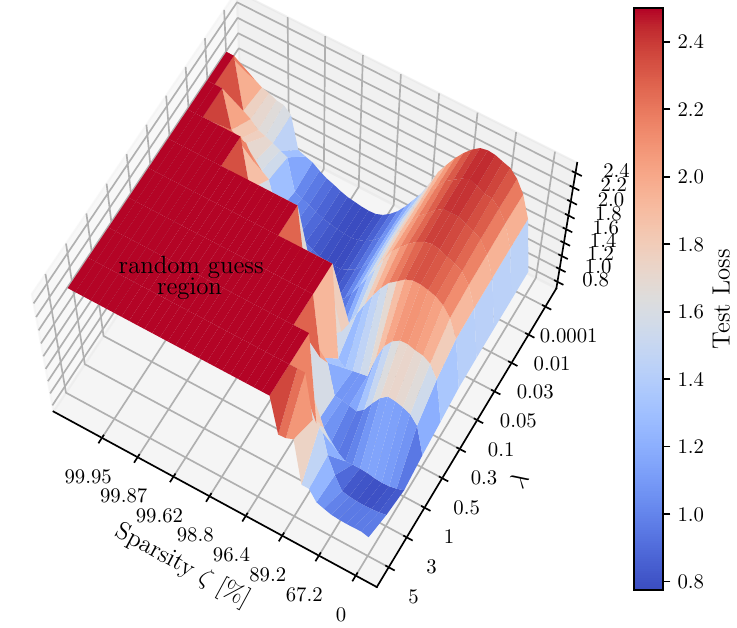}
    \caption{Ablation study over $\lambda$: test loss of ViT on CIFAR-10 with $\varepsilon = 10\%$.}
    \label{fig:Varying lambda for CIFAR-10}
\end{figure}
\textbf{Setup.} For the experimental setup, we follow the same approach as He~et~al.~\cite{SparseDoubleDescent}. The first model we train is a ResNet-18, trained on CIFAR-10 \& CIFAR-100, for 160 epochs, optimized with SGD, having momentum 0.9, a learning rate of 0.1 decayed by a factor 0.1 at milestones 80 and 120, batch size 128 and $\lambda$ $10^{-4}$. The second model is a ViT with 4 patches, 8 heads, and 512 embedding dimensions, trained on CIFAR-10 and CIFAR-100 for 200 epochs, optimized with Adam, having a learning rate of $10^{-4}$ with a cosine annealing schedule and $\lambda$ 0.03.
For each dataset, a percentage $\varepsilon$ of symmetric, noisy labels are introduced: the labels of a given proportion of training samples are flipped to one of the other class labels, selected with equal probability~\cite{Noisy_labels}. In our experiments, we test with $\varepsilon \in \{10\%, 20\%, 50\%\}$. Moreover, as synthetic noise has clean structures which greatly enabled statistical analyses but often fails to model real-world noise patterns, we also conducted experiments without adding synthetic noise. With the same architectures and learning policies presented above, we carried out experiments on CIFAR-100N, which is formed by the CIFAR-100 training dataset with human-annotated real-world noisy labels collected from Amazon Mechanical Turk~\cite{wei2022learning}. 
In all experiments, we set $\zeta^{\text{iter}}=20\%$ and $\zeta^{\text{end}}=99.99\%$.\footnote{The code is available at \url{https://github.com/VGCQ/SDD_ViT}}

\noindent\textbf{Occurrence of sparse double descent.} Fig.~\ref{fig:CIFAR-ViT} displays the results of ResNet-18 and ViT, on CIFAR-10 and CIFAR-100. As in He~et~al.~\cite{SparseDoubleDescent} work, the double descent consists of 4 phases. First, at low sparsities, the network is overparameterized, thus pruned network can still reach similar accuracy to the dense model. The second phase is a phase near the ``interpolation threshold'', where 
the test accuracy is about to first decrease and then increase as sparsity grows. The third phase is located at high sparsities, where test accuracy is rising. The final phase happens when both training and test accuracy drop significantly. For every value of $\varepsilon$, whether on CIFAR-10 or CIFAR-100, the sparse double descent phenomenon occurs both for ResNet and ViT. We observe a similar phenomenon as in the simulated $\varepsilon$ also in the human-annotated CIFAR-100N.

\noindent\textbf{Study on $\boldsymbol{\lambda}$.} In the previous experiments in Fig.~\ref{fig:CIFAR-ViT}, ViTs were trained with a $\ell_2$-regularization hyper-parameter equal to 0.03, which is typically used in other works.
However, it has been recently shown that, for certain linear regression models with isotropic data distribution, optimally-tuned $\ell_2$ regularization can achieve monotonic test performance as either the sample size or the model size is grown. Nakkiran~et~al.~\cite{nakkiran2021optimal} demonstrated it analytically and established that optimally-tuned $\ell_2$ regularization can mitigate double descent for general models, including neural networks like Convolutional Neural Networks. Moreover, a recent study showed that $\ell_2$ regularization is positively contributing to the avoidance of sparse double descent in an image classification context, but is not the antidote to ``dodge'' it~\cite{quétu2023avoid}. 
Hence, we propose in Fig.~\ref{fig:Varying lambda for CIFAR-10} a quantitative study over $\lambda$ for ViT on CIFAR-10 with $\varepsilon=10\%$. 
With small values of $\lambda$, i.e. below $1$, the sparse double descent is empirically noticeable. The increment of $\lambda$ pushes the occurrence of the phenomenon towards smaller sparsity values. Looking at the loss, increasing $\lambda$ smoothens the bump of the test loss and at some point, i.e. $\lambda=1$, the test loss becomes flat and behaves monotonically: the sparse double descent is avoided. For $\lambda>1$, the phenomenon also results avoided, but the performance worsens (lighter blue region at the bottom right corner) since the regularization is stronger. Note that with higher $\lambda$, performances are better but the maximum sparsity achievable is not as high as for lower values of $\lambda$.
\begin{figure*}[t]
    \begin{subfigure}{0.5\textwidth}
        \includegraphics[width=\textwidth]{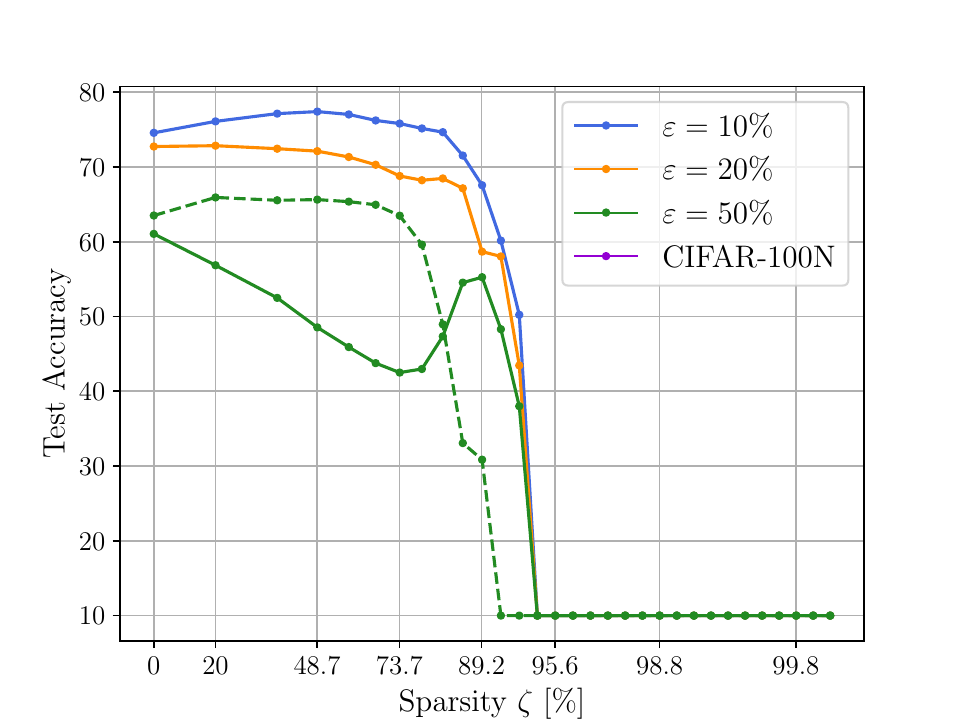}
        \caption{ViT on CIFAR-10}
    \end{subfigure}
    \begin{subfigure}{0.5\textwidth}
        \includegraphics[width=\textwidth]{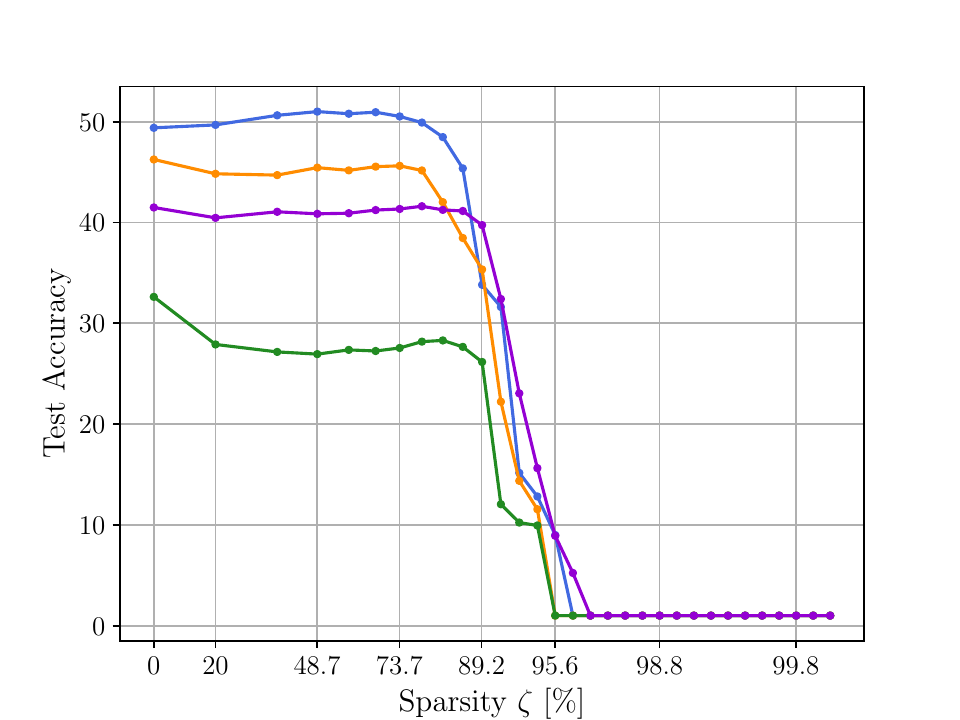}
        \caption{ViT on CIFAR-100}
    \end{subfigure}
    \caption{Test accuracy of ViT on CIFAR dataset with different amount of noise $\varepsilon$ with $\lambda=1$ (solid lines), $\lambda=3$ (dashed line).}
    \label{fig:CIFAR-ViT-wd}
\end{figure*}
\begin{figure*}[t]
    \begin{subfigure}{0.5\textwidth}
        \includegraphics[width=\textwidth]{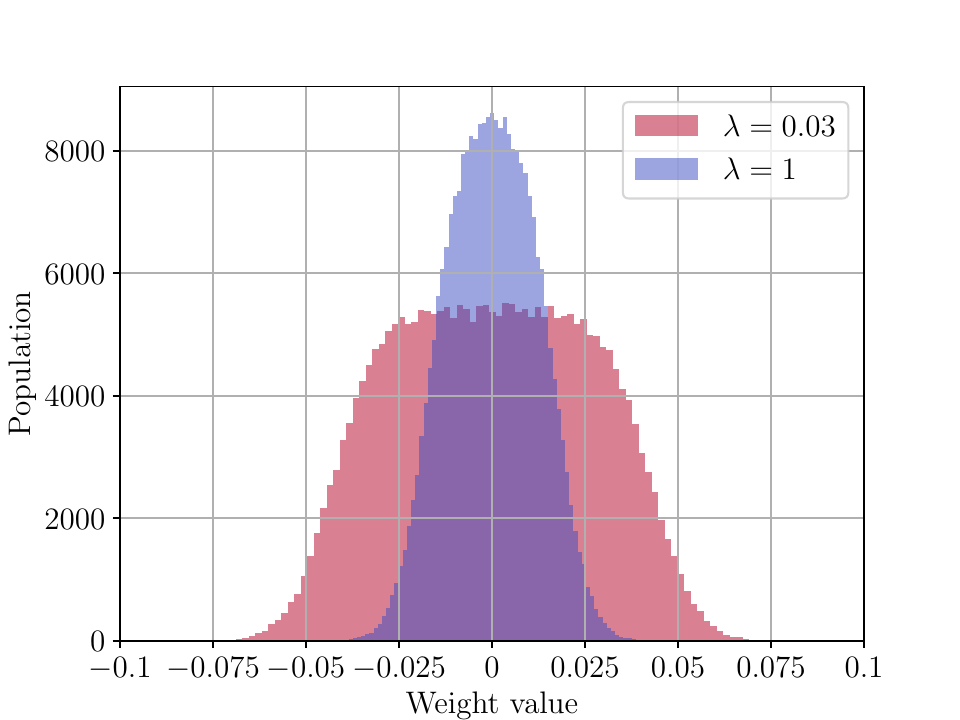}
        \caption{~}
    \end{subfigure}
    \begin{subfigure}{0.5\textwidth}
        \includegraphics[width=\textwidth]{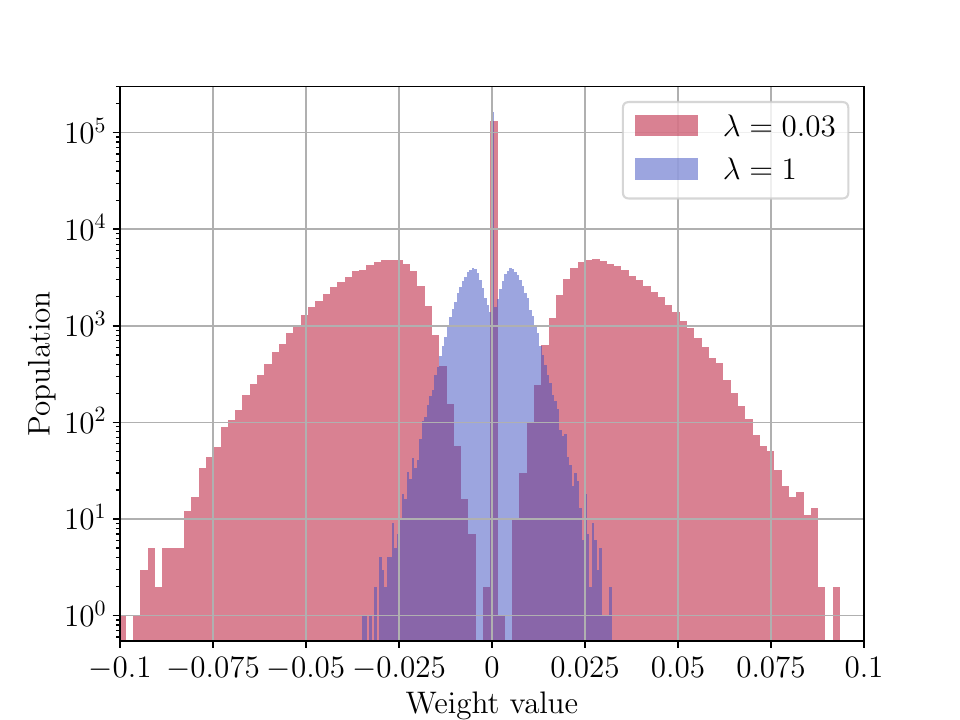}
        \caption{~}
    \end{subfigure}
    \caption{Histogram of the weights of ViT for $\varepsilon=10\%$ on CIFAR-10,\\ with $\zeta\!=\!0\%$~(a) and $\zeta=48.8\%$~(b).}
    \label{fig:Histograms}
\end{figure*}

\noindent\textbf{Avoidance of the Sparse Double Descent.} 
As $\lambda=1$ seems to be an optimal value enabling dodging SDD on CIFAR-10 with $\varepsilon=10\%$, we try to use this value for other setups. Fig.~\ref{fig:CIFAR-ViT-wd} displays the results of ViT on CIFAR-10/CIFAR-100 with $\varepsilon \in \{10\%, 20\%, 50\%\}$ and $\lambda=1$. For small noise rates, i.e. $\varepsilon \leq 20\%$, the phenomenon vanishes and performance is enhanced. However, for higher noise rates, like $\varepsilon=50\%$, SDD is mitigated, but still present. Even if it already helps, it seems that the strength of the regularization is not high enough to completely avoid SDD. Indeed, with a higher $\lambda$, i.e. 3, the performance becomes monotonic.
\noindent\textbf{Trade-off between SDD and compressibility.} Fig.~\ref{fig:Varying lambda for CIFAR-10}, supported also by the experiments displayed in Fig.~\ref{fig:CIFAR-ViT-wd}, suggests that at high regularization regimes, where we avoid SDD, the ability to compress the model is harmed. This is due to the strong prior we impose over the distribution of the parameters of the model: the stronger this is, the least we are indeed able to remove degrees of freedom from our system. As a visual example, Fig.~\ref{fig:Histograms} displays the distribution of the parameters for one of the considered training configurations, for $\lambda=0.03$ and $1$, without pruning and after two pruning steps. We observe that despite removing the same quantity of parameters, with higher regularization the parameters have less variance, which has the dual effect of both making them more robust to injected noise (due to the strong regularization) but, at the same time, this distribution is more sensitive to compression by pruning. Hence, we conclude that, in case we wish to have a robust, well-generalizing model, we wish to avoid SDD and employ strong $\ell_2$ regularization; on the contrary, if we target compressibility, we would like to favor SDD, as the better generalizing region is pushed to highly compressed regions.
\section{Conclusion}

This paper investigates the occurrence of Sparse Double Descent in the Vision Transformer architecture. SDD is a phenomenon carefully explored due to its influence on determining the optimal model size necessary for maintaining the performance of over-parametrized models. We observe that, indeed, ViT is also susceptible to SDD. Moreover, we study different values for $\ell_2$ regularization and discover that, unlike for other CNN architectures like ResNet, we can find the optimal value and completely avoid SDD. However, the regularization comes at a price - at the same time, it renders the model less compressible, because of the strong enforced prior. We postulate that this is possible due to the lack of strong inductive bias in ViT, which enables strong regularization regimes, impossible for CNNs. Finally, we inspect the trade-off between avoiding SDD (enhancing hence model's performance) and favoring the model compressibility, observing that, for the second one, we would like to favor SDD. This study hopes to inform the community about the risk of SDD ViT models might incur, which depending on the final scope of the trained model can be a real or a phantom threat.

\noindent\textbf{Acknowledgments.} This project was provided with computer and storage resources by GENCI at IDRIS thanks to the grant 2022-AD011013930 on the supercomputer Jean Zay's the V100 partition.
%
%
%
\bibliographystyle{splncs04}
\bibliography{main}

\begin{thebibliography}{10}
\providecommand{\url}[1]{\texttt{#1}}
\providecommand{\urlprefix}{URL }
\providecommand{\doi}[1]{https://doi.org/#1}

\bibitem{barbano2022two}
Barbano, C.A., Tartaglione, E., Berzovini, C., Calandri, M., Grangetto, M.: A
  two-step radiologist-like approach for covid-19 computer-aided diagnosis from
  chest x-ray images. In: Image Analysis and Processing--ICIAP 2022: 21st
  International Conference, Lecce, Italy, May 23--27, 2022, Proceedings, Part
  I. pp. 173--184. Springer (2022)

\bibitem{brown2020language}
Brown, T., Mann, B., Ryder, N., Subbiah, M., Kaplan, J.D., Dhariwal, P.,
  Neelakantan, A., Shyam, P., Sastry, G., Askell, A., et~al.: Language models
  are few-shot learners. Advances in neural information processing systems
  \textbf{33},  1877--1901 (2020)

\bibitem{chaudhry2022lung}
Chaudhry, H.A.H., Renzulli, R., Perlo, D., Santinelli, F., Tibaldi, S.,
  Cristiano, C., Grosso, M., Fiandrotti, A., Lucenteforte, M., Cavagnino, D.:
  Lung nodules segmentation with deephealth toolkit. In: Image Analysis and
  Processing. ICIAP 2022 Workshops: ICIAP International Workshops, Lecce,
  Italy, May 23--27, 2022, Revised Selected Papers, Part I. pp. 487--497.
  Springer (2022)

\bibitem{dai2021coatnet}
Dai, Z., Liu, H., Le, Q.V., Tan, M.: Coatnet: Marrying convolution and
  attention for all data sizes. Advances in Neural Information Processing
  Systems  \textbf{34},  3965--3977 (2021)

\bibitem{dehghani2023scaling}
Dehghani, M., Djolonga, J., Mustafa, B., Padlewski, P., Heek, J., Gilmer, J.,
  Steiner, A., Caron, M., Geirhos, R., Alabdulmohsin, I., et~al.: Scaling
  vision transformers to 22 billion parameters. arXiv preprint arXiv:2302.05442
   (2023)

\bibitem{dosovitskiy2021an}
Dosovitskiy, A., Beyer, L., Kolesnikov, A., Weissenborn, D., Zhai, X.,
  Unterthiner, T., Dehghani, M., Minderer, M., Heigold, G., Gelly, S.,
  Uszkoreit, J., Houlsby, N.: An image is worth 16x16 words: Transformers for
  image recognition at scale. In: International Conference on Learning
  Representations (2021)

\bibitem{DosovitskiyB0WZ21}
Dosovitskiy, A., Beyer, L., Kolesnikov, A., Weissenborn, D., Zhai, X.,
  Unterthiner, T., Dehghani, M., Minderer, M., Heigold, G., Gelly, S.,
  Uszkoreit, J., Houlsby, N.: An image is worth 16x16 words: Transformers for
  image recognition at scale. In: 9th International Conference on Learning
  Representations, {ICLR} 2021, Virtual Event, Austria, May 3-7, 2021.
  OpenReview.net (2021)

\bibitem{esser2021taming}
Esser, P., Rombach, R., Ommer, B.: Taming transformers for high-resolution
  image synthesis. In: Proceedings of the IEEE/CVF conference on computer
  vision and pattern recognition. pp. 12873--12883 (2021)

\bibitem{Gale_Magnitude}
Gale, T., Elsen, E., Hooker, S.: The state of sparsity in deep neural networks.
  arXiv preprint arXiv:1902.09574  (2019)

\bibitem{han2015learning}
Han, S., Pool, J., Tran, J., Dally, W.: Learning both weights and connections
  for efficient neural network. Advances in neural information processing
  systems  \textbf{28} (2015)

\bibitem{SparseDoubleDescent}
He, Z., Xie, Z., Zhu, Q., Qin, Z.: Sparse double descent: Where network pruning
  aggravates overfitting. In: International Conference on Machine Learning. pp.
  8635--8659. PMLR (2022)

\bibitem{transformers_in_vision_survey2022}
Khan, S., Naseer, M., Hayat, M., Zamir, S.W., Khan, F.S., Shah, M.:
  Transformers in vision: A survey. ACM Comput. Surv.  \textbf{54}(10s) (sep
  2022)

\bibitem{Liu2021SwinTH}
Liu, Z., Lin, Y., Cao, Y., Hu, H., Wei, Y., Zhang, Z., Lin, S., Guo, B.: Swin
  transformer: Hierarchical vision transformer using shifted windows. 2021
  IEEE/CVF International Conference on Computer Vision (ICCV) pp. 9992--10002
  (2021)

\bibitem{ma2020normalized}
Ma, X., Huang, H., Wang, Y., Romano, S., Erfani, S., Bailey, J.: Normalized
  loss functions for deep learning with noisy labels. In: International
  conference on machine learning. pp. 6543--6553. PMLR (2020)

\bibitem{Noisy_labels}
Ma, X., Wang, Y., Houle, M.E., Zhou, S., Erfani, S., Xia, S., Wijewickrema, S.,
  Bailey, J.: Dimensionality-driven learning with noisy labels. In:
  International Conference on Machine Learning. pp. 3355--3364. PMLR (2018)

\bibitem{mazzeo2022image}
Mazzeo, P.L., Frontoni, E., Sclaroff, S., Distante, C.: Image Analysis and
  Processing. ICIAP 2022 Workshops: ICIAP International Workshops, Lecce,
  Italy, May 23--27, 2022, Revised Selected Papers, Part I, vol. 13373.
  Springer Nature (2022)

\bibitem{Nakkiran2021Deep}
Nakkiran, P., Kaplun, G., Bansal, Y., Yang, T., Barak, B., Sutskever, I.: Deep
  double descent: Where bigger models and more data hurt. In: International
  Conference on Learning Representations (2020)

\bibitem{nakkiran2021optimal}
Nakkiran, P., Venkat, P., Kakade, S.M., Ma, T.: Optimal regularization can
  mitigate double descent. In: International Conference on Learning
  Representations (2021)

\bibitem{quetu2023dodging}
Qu{\'e}tu, V., Tartaglione, E.: Dodging the sparse double descent. arXiv
  preprint arXiv:2303.01213  (2023)

\bibitem{quétu2023avoid}
Quétu, V., Tartaglione, E.: Can we avoid double descent in deep neural
  networks? (2023)

\bibitem{SpringenbergDBR14}
Springenberg, J.T., Dosovitskiy, A., Brox, T., Riedmiller, M.A.: Striving for
  simplicity: The all convolutional net. In: Bengio, Y., LeCun, Y. (eds.) 3rd
  International Conference on Learning Representations, {ICLR} 2015, San Diego,
  CA, USA, May 7-9, 2015, Workshop Track Proceedings (2015)

\bibitem{sukhbaatar2014training}
Sukhbaatar, S., Bruna, J., Paluri, M., Bourdev, L., Fergus, R.: Training
  convolutional networks with noisy labels. arXiv preprint arXiv:1406.2080
  (2014)

\bibitem{touvron2021training}
Touvron, H., Cord, M., Douze, M., Massa, F., Sablayrolles, A., J{\'e}gou, H.:
  Training data-efficient image transformers \& distillation through attention.
  In: International conference on machine learning. pp. 10347--10357. PMLR
  (2021)

\bibitem{vaswani2017attention}
Vaswani, A., Shazeer, N., Parmar, N., Uszkoreit, J., Jones, L., Gomez, A.N.,
  Kaiser, {\L}., Polosukhin, I.: Attention is all you need. Advances in neural
  information processing systems  \textbf{30} (2017)

\bibitem{wei2022learning}
Wei, J., Zhu, Z., Cheng, H., Liu, T., Niu, G., Liu, Y.: Learning with noisy
  labels revisited: A study using real-world human annotations. In:
  International Conference on Learning Representations (2022)

\bibitem{yilmaz2022regularization}
Yilmaz, F.F., Heckel, R.: Regularization-wise double descent: Why it occurs and
  how to eliminate it. In: 2022 IEEE International Symposium on Information
  Theory (ISIT). pp. 426--431. IEEE (2022)

\bibitem{yu2022width}
Yu, F., Huang, K., Wang, M., Cheng, Y., Chu, W., Cui, L.: Width \& depth
  pruning for vision transformers. In: Proceedings of the AAAI Conference on
  Artificial Intelligence. vol.~36, pp. 3143--3151 (2022)

\bibitem{yuan2021tokens}
Yuan, L., Chen, Y., Wang, T., Yu, W., Shi, Y., Jiang, Z.H., Tay, F.E., Feng,
  J., Yan, S.: Tokens-to-token vit: Training vision transformers from scratch
  on imagenet. In: Proceedings of the IEEE/CVF international conference on
  computer vision. pp. 558--567 (2021)

\end{thebibliography}
\end{document}